\documentclass{article}

\usepackage{arxiv}

\usepackage{times}
\usepackage{epsfig}
\usepackage{graphicx}
\usepackage{subfigure}
\usepackage{amsmath}
\usepackage{amssymb}

\begin{document}

\title{MimosaNet: An Unrobust Neural Network Preventing Model Stealing}
\author{K\'alm\'an Szentannai, Jalal Al-Afandi, Andr\'as Horv\'ath\\
Peter Pazmany Catholic University - Faculty of Information Technology and Bionics\\
Pr\'ater u. 50/A, 1083, Budapest, Hungary\\
\tt\small szentannai.kalman@hallgato.ppke.hu, alafandi.mohammad.jalal@itk.ppke.hu,\\\tt\small horvath.andras@itk.ppke.hu
}

\maketitle

\begin{abstract}

Deep Neural Networks are robust to minor perturbations of the learned network parameters and their minor modifications do not change the overall network response significantly.
This allows space for model stealing, where a malevolent attacker can steal an already trained network, modify the weights and claim the new network his own intellectual property. In certain cases this can prevent the free distribution and application of networks in the embedded domain.
In this paper, we propose a method for creating an equivalent version of an already trained fully connected deep neural network that can prevent network stealing: namely, it produces the same responses and classification accuracy, but it is extremely sensitive to weight changes. 
\end{abstract}

\section{Introduction}\label{Intro}
Deep neural networks are employed in an emerging number of tasks, many of which were not solvable before with traditional machine learning approaches. 
In these structures, expert knowledge which is represented in annotated datasets is transformed into learned network parameters known as network weights during training. 

Methods, approaches and network architectures are distributed openly in this community, but most companies 
protect their data and trained networks obtained from tremendous amount of working hours annotating datasets and fine-tuning training parameters.

Model stealing and detection of unauthorized use via stolen weights is a key challenge of the field as there are techniques (scaling, noising, fine-tuning, distillation) to modify the weights to hide the abuse, while preserving the functionality and accuracy of the original network. Since networks are trained by stochastic optimization methods and are initialized with random weights, training on a dataset might result various different networks with similar accuracy. 

There are several existing methods to measure distances between network weights after these modifications and independent trainings: \cite{koch1995towards} \cite{wolfgang1996watermark} \cite{zarrabi2018reversible}
Obfuscation of neural networks was introduced in \cite{xu2018deepobfuscation}, which showed the viability and importance of these approaches. In this paper the authors present a method to obfuscate the architecture, but not the learned network functionality. We would argue that most ownership concerns are not raised because of network architectures, since most industrial applications use previously published structures, but because of network functionality and the learned weights of the network.
%

Other approaches try to embed additional, hidden information in the network such as hidden functionalities or non-plausible, predefined answers for previously selected images (usually referred as watermarks) \cite{namba2019robust}, \cite{gomez2018intellectual}. In case of a stolen network one can claim ownership of the network by unraveling the hidden functionality, which can not just be formed randomly in the structure.
A good summary comparing different watermarking methods and their possible evasions can be found in \cite{hitaj2018have}.

Instead of creating evidence, based on which relation between the original and the stolen, modified model could be proven, we have developed a method which generates a completely sensitive and fragile network, which can be freely shared, since even minor modification of the network weights would drastically alter the networks response.

In this paper 
we present a method which can transform a previously trained network into a fragile one, by extending the number of neurons in the selected layers, without changing the response of the network. These transformations can be applied in an iterative manner on any layer of the network, expect the first and the last layers (since their size is determined by the problem representation).

\section{Mathematical model of Unrobust Networks}\label{Transform}
\subsection{Fully Connected Layers}\label{mathematical_model}

In this section we would like to present our method, how a robust network can be transformed into a non-robust one. We have chosen fully connected networks because of their generality and compact mathematical representation.
Fully connected networks are generally applied covering the whole spectrum of machine learning problems from regression through data generation to classification problems. The authors can not deny the fact, that in most practical problems convolutional networks are used, but we would like to emphasize the following properties of fully connected networks:
\textbf{(1)} in those cases when there is no topographic correlation in the data fully connected networks are applied \textbf{(2)} most problems also contain additional fully connected layers after the feature extraction of the convolutional or residual layers \textbf{(3)} every convolutional network can be considered a special case of fully connected ones, where all weights outside the convolutional kernels are set to zero.

A fully connected deep neural network might consist of several hidden layers each containing certain number of neurons. Since all layers have the same architecture, without the loss of generality, we will focus here only on three consecutive layers in the network ($i-1$, $i$ and $i+1$).
We will show how neurons in layer $i$ can be changed, increasing the number of neurons in this layer and making the network fragile, meanwhile keeping the functionality of the three layers intact.
We have to emphasize that this method can be applied on any three layers, including the first and last three layers of the network and also that it can be applied repeatedly on each layer, still without changing the overall functionality of the network.

The input of the layer $i$, the activations of the previous layer ($i-1$) can be noted by the vector $x_{i-1}$ containing $N$ elements.
The weights of the network are noted by the weight matrix $W_i$ and the bias $b_i$ where W is a matrix of $N \times K$ elements, creating a mapping $\mathbb{R}^N \mapsto  \mathbb{R}^K$ and $b_i$ is a vector containing $K$ elements.
The output of layer $i$, also the input of layer $i+1$ can be written as:

\begin{equation}
\label{eq:basic}
x_i=\phi(W_{i_{N \times K}} x_{i-1} + b_i)
\end{equation}

where $\phi$ is the activation function of the neurons.

The activations of layer $i+1$ can be extended as using equation \ref{eq:basic}:

\begin{equation}
\label{eq:three}
x_{i+1} = \phi(\phi(x W_{i-1_{N \times K}}  + b_{i-1})W_{i_{K \times L}} + b_i)
\end{equation}
Creating a mapping $\mathbb{R}^N \mapsto  \mathbb{R}^L$. 

One way of identifying a fully connected neural network is to represent it as a sequence of synaptic weights. Our assumption was that in case of model stealing certain application of additive noise on the weights would prevent others to reveal the attacker and conceal thievery. Since fully connected networks are known to be robust against such modifications, the attacker could use the modified network with approximately the same classification accuracy. 
Thus, our goal was to find a transformation that preserves the loss and accuracy rates of the network, but introduces a significant decrease in terms of the robustness against parameter tuning. In case of a three-layered structure one has to preserve the mapping between the first and third layers (EQ \ref{eq:three}) to keep the functionality of this three consecutive layers, but the mapping in EQ \ref{eq:basic} (the mapping between the first and second, or second and third layers), can be changed freely.

Also, our model must rely on an identification mechanism based on a representation of the synaptic weights. Therefore, the owner of a network should be able to verify the ownership based on the representation of the neural network, examining the average distance between the weights\cite{hitaj2018have}.

\subsection{Decomposing Neurons}\label{descomposition}

We would like to find such $W'_{i-1_{N \times M}}$ and $W'_{i_{M \times L}}$ ($M \in \mathbb{N} , M > K$) matrices, for which:

\begin{equation}\label{nonlineq}
\begin{split}
\phi(\phi(x W_{i-1_{N \times K}} + b_{i-1}) W_{i_{K\times L}} + b_i) \\
= \phi(\phi(x W'_{i-1_{N \times M}} + b'_{i-1}) W'_{i_{M\times L}} + b_i)
\end{split}
\end{equation}

Considering the linear case when \(\phi(x) = x\) we obtain the following form:

\begin{equation}\label{lineq}
\begin{split}
x W_{i-1_{N \times K}} W_{i_{K\times L}} + b_{i-1} W_{i_{K\times L}} + b_i \\
= x W'_{i-1_{N \times M}} W'_{i_{M\times L}} + b'_{i-1} W'_{i_{M\times L}} + b_i
\end{split}
\end{equation}

The equation above holds only for the special case of $\phi(x)=x$, however in most cases nonlinear activation functions are used. We have selected the rectified linear unit (ReLU) for our investigation ($\phi(x) = max(0,x)$). This non-linearity consist of two linear parts, which means that a variable could be in a linear domain of EQ \ref{nonlineq} resulting selected lines of \ref{lineq} (if $x\geq0$), or the equation system is independent from the variable if the activation function results a constant zero (if $x\leq0$). This way ReLU gives a selection of given variables (lines) of \ref{lineq}. However, applying the ReLU activation function has certain constraints.

Assume, that a neuron with the ReLU activation function should be replaced by two other neurons. This can be achieved by using an $\alpha \in (0,1)$ multiplier:

\begin{equation}
\phi(\sum_{i=1}^{n}W_{ji}^l x_i + b_j^l) = N_j^l
\end{equation}

\begin{equation}
N_j^l = \alpha N_j^l + (1-\alpha) N_j^l
\end{equation}
where $\alpha N_j^l$ and $(1-\alpha) N_j^l$ correspond to the activation of the two neurons. For each of these, the activation would only be positive if the original neuron had a positive activation, otherwise it would be zero, this means that all the decomposed neuron must have the same bias.

After decomposing a neuron, it is needed to choose the appropriate weights on the subsequent layer. A trivial solution is to keep the original synaptic weights represented by the $\overline{W}_{j}^{l+1}$ column vector. This would lead to the same activation since 
\begin{equation} \label{alphadecomposition}
N_j^l \overline{W}_{j}^{l+1} = \alpha N_j^l \overline{W}_{j}^{l+1} + (1-\alpha) N_j^l \overline{W}_{j}^{l+1}
\end{equation}

A fragile network can be created by choosing the same synaptic weights for the selected two neurons, but it would be easy to spot by the attacker, thus another solution is needed. In order to find a nontrivial solution we constructed a linear equation system that can be described by equation system $A p = c$, where $A$ contains the original, already decomposed synaptic weights of the first layer, meanwhile, $p$ represents the unknown synaptic weights of the subsequent layer. Vector $c$ contains the corresponding weights from the original network multiplied together: each element represents the amount of activation related to one input. Finally the non-trivial solution can be obtained by solving the following non-homogeneous linear equation system for each output neuron where index $j$ denotes the output neuron. 
\small
\begin{equation} \label{nontrivial}
\begin{bmatrix}
w_{11}^{1'} & w_{21}^{1'} & \hdots & w_{m1}^{1'} \\
w_{12}^{1'} & w_{22}^{1'} & \hdots & w_{m2}^{1'} \\
\vdots & \vdots & \ddots & \vdots \\
w_{1n}^{1'} & w_{2n}^{1'} & \hdots & w_{mn}^{1'} \\
b_{1}^{1'} & b_{2}^{1'} & \hdots & b_{m}^{1'}
\end{bmatrix}
\times
\begin{bmatrix}
w_{j1}^{2'}\\
w_{j2}^{2'}\\
\vdots \\
w_{jm}^{2'}\\
\end{bmatrix}
=
\begin{bmatrix}
\sum_{i=1}^{k}w_{ji}^2 w_{i1}^1\\
\sum_{i=1}^{k}w_{ji}^2 w_{i2}^1\\
\vdots \\
\sum_{i=1}^{k}w_{ji}^2 w_{ik}^1\\
\sum_{i=1}^{k}w_{ji}^2 b_{i}^1
\end{bmatrix}
\end{equation}
\normalsize
It is important to note, that all the predefined weights on layer $l+1$ might change. In summary, this step can be considered as the replacement of a layer, changing all synaptic weights connecting from and to this layer, but keeping the biases of the neurons and the functionality of the network intact.

The only constraint of this method is related to the number of neurons regarding the two consecutive layers. It is known, that for matrix $A$ with the size of $M \times N$, equation $A p = c$  has a solution if and only if $rank(A) = rank[A|c]$ where $[A|c]$ is the extended matrix. 
The decomposition of a neuron described in equation \ref{alphadecomposition} results in linearly dependent weight vectors on layer $l$, therefore when solving the equation system the rank of the matrix $A$ is less than or equal to $min(N+1,K)$. If the rank is equal to $N+1$ (meaning that $K \geq N+1$) then vector $c$ with the dimension of $N+1$ would not introduce a new dimension to the subspace defined by matrix $A$. However if $rank(A) = K$ (meaning that $K \leq N+1$) then vector $c$ could extend the subspace defined by $A$. Therefore, the general condition for solving the equation system is: $K \geq N+1$.

This shows that one could increase the number of the neurons in a layer, and divide the weights of the existing neuron in that layer.  We have used this derivation and aim to find a solution of EQ. \ref{alphadecomposition} where the order of magnitudes are significantly different (in the range of $10^6$) for both the network parameters and for the eigenvalues of the mapping  $\mathbb{R}^N \mapsto  \mathbb{R}^L$.

\subsection{Introducing Deceptive Neurons}

The method described in the previous section results a fragile neural network, but unfortunately it is not enough to protect the network weights, since an attacker could identify the decomposed neurons based on their biases or could fit a new neural network on the functionality implemented by the layer. To prevent this we will introduce deceptive neurons in layers.
The purpose of these neurons is to have non-zero activation in sum if and only if noise was added to their weights apart from this all these neurons have to cancel each others effect out in the network, but not necessarily in a single layer.

The simplest method is to define a neuron with an arbitrary weight and a bias of an existing neuron resulting a large activation and making a copy of it with the difference of multiplying the output weights by $-1$. As a result, these neurons do not contribute to the functionality of the network. However, adding noise to the weights of these neurons would have unexpected consequences depending on the characteristics of the noise, eventually leading to a decrease of classification accuracy. 

One important aspect of this method is to hide the generated neurons and obfuscate the network to prevent the attacker to easily filter our deceptive neurons in the architecture. Choosing the same weights again on both layers would be an obvious sign to an attacker, therefore this method should be combined with decomposition described in \ref{descomposition}. 

Since decomposition allows the generation of arbitrarily small weights one can select a suitably small magnitude, which allows the generation of $R$ real (non deceptive) neurons in the system, and half of their weights ($\alpha$ parameters) can be set arbitrarily, meanwhile the other half of the weights will be determined by EQ. \ref{nontrivial}. For each real neuron one can generate a number ($F$) of fake neurons forming groups of $R$ number of real and $F$ number of fake neurons. These groups can be easily identified in the network since all of them will have the same bias, but the identification of fake and real neurons in a group is non-polynomial.

The efficiency of this method should be measured in the computational complexity of successfully finding two or more corresponding fake neurons having a total activation of zero in a group. Assuming that only one pair of fake neurons was added to the network, it requires $\sum_{i=0}^{L}{R_{i}+F_{i} \choose 2}$ steps to successfully identify the fake neurons, where $R_{i}+F_{i}$ denotes the number of neurons in the corresponding hidden layer, and $L$ is the number of hidden layers. This can be further increased by decomposing the fake neurons using EQ. \ref{nontrivial}: in that case the required number of steps is $\sum_{i=0}^{L}{R_{i}+F_{i} \choose d+2}$, $d$ being the number of extra decomposed neurons. This can be maximized if $d+2=R_{i}+F_{i}/2$, where $i$ denotes the layer, where the fake neurons are located. However, this is true only if the attacker has information about the number of deceptive neurons. Without any prior knowledge, the attacker has to guess the number of deceptive neurons  as well ($0,1,2 \dots {R_{i}+F_{i}-1}$) which leads to exponentially increasing computational time.

\section{Experiments}\label{Experiments}
\subsection{Simulation of a Simple Network}
As a case study we have created a simple fully connected neural network with three layers, each containing two neurons to present the validity of our approach. The functionality of the network can be considered as a mapping  $f: \mathbb{R}^2 \mapsto \mathbb{R}^2$.

\vspace{-7mm}
\footnotesize
\begin{center}
\begin{equation*}\label{simple_network}
w_1 =
\begin{bmatrix}
6 & -1 \\
-1 & 7 \\
\end{bmatrix}
,\
b_1 =
\begin{bmatrix}
1 & -5 \\
\end{bmatrix}
w_2 =
\begin{bmatrix}
5 & 3 \\
9 & -1 \\
\end{bmatrix}
,\
b_2 =
\begin{bmatrix}
7 & 1 \\
\end{bmatrix}
\end{equation*}
\end{center}

\normalsize
We added two neurons to the hidden layer with decomposition, which does not modify the input and output space and no deceptive neurons were used in this experiment. After applying the methods described in \ref{mathematical_model} we obtained a solution of:

\scriptsize
\begin{center}
\begin{equation*}\label{simple_network}
\begin{split}
w_1 =
\begin{bmatrix}
0.0525 & -0.4213 & 6.0058 &  -0.5744\\
-0.0087 & 2.9688 & -0.9991 & 4.0263\\
\end{bmatrix}
\\
b_1 =
\begin{bmatrix}
0.0087 & -2.1066 & 1.0009 & -2.8722 \\
\end{bmatrix}
\\
w_2 =
\begin{bmatrix}
4.1924e+03 & -5.4065e+03\\
-2.3914 & 7.3381 \\
-3.2266 & 5.7622 \\
6.9634 & -7.0666 \\
\end{bmatrix}
\\
b_2 =
\begin{bmatrix}
7 & 1 \\
\end{bmatrix}
\end{split}
\end{equation*}
\end{center}

\begin{figure}[!htp]
\centering
\subfigure{\includegraphics[width=3.6in,height=1.8in]{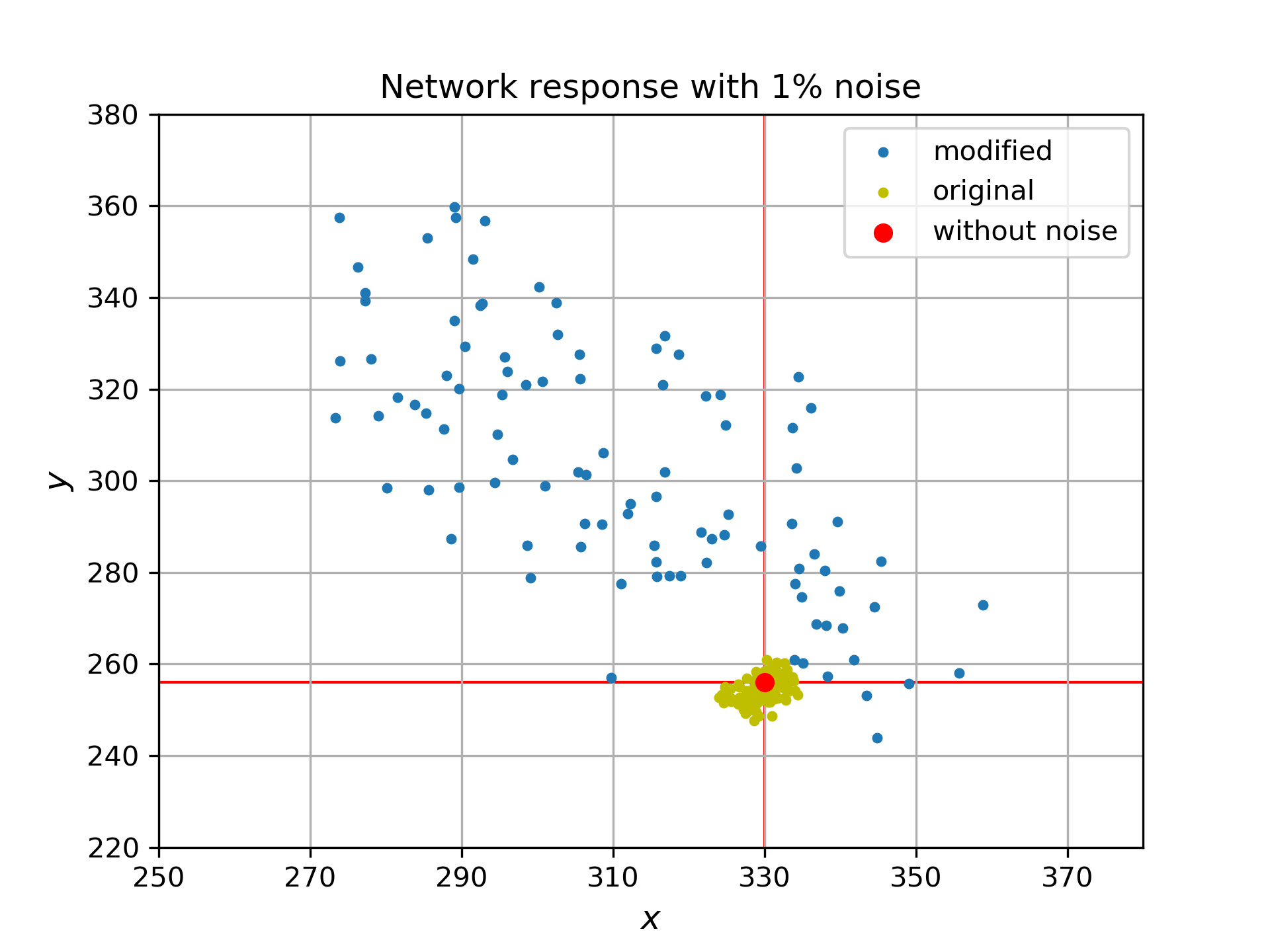}}
\caption{This figure depicts the response of a simple two-layered fully connected network for a selected input (red dot) and the response of its variants with \%1 noise (yellow dots) added proportionally to the weights. The blue dots represent the responses of the transformed MimosaNets under the same level of noise on their weights, meanwhile the response of the transformed network (without noise) remained exactly the same.  }
\label{mixed_reponse}
\end{figure}

\normalsize
In the following experiment we have chosen an arbitrary input vector: $[7,9]$. We have measured the response of the network for this input, each time introducing 1\% noise to the weights of the network. Figure \ref{mixed_reponse} shows the response of the original network and the modified network after adding 1\% noise. The variances of the original network for the first output dimension is 6.083 and 8.399 for the second, meanwhile the variances are 476.221 and 767.877 for the decomposed networks respectively. This example demonstrates how decomposition of a layer can increase the networks dependence on its weights.

\subsection{Simulations on the MNIST Dataset}

We have created a five layered fully connected network containing 32 neurons in each hidden-layer (and 728 and 10 neurons in the input and output layers) and trained it on the MNIST\cite{lecun2010mnist} dataset, using batches of 32 and Adam Optimizer\cite{kingma2014adam} for 7500 iterations. The network has reached an accuracy of $98.4\%$ on the independent test set.

We have created different modifications of the network by adding 9,18,36,72 extra neurons. These neurons were divided equally between the three hidden-layers and 2/3 of them were deceptive neurons (since they were always created in pairs) and 1/3 of them were created by decomposition. This means that in case of 36 additional neurons $2\times4$ deceptive neurons were added to each layer and four new neurons per layer were created by decomposition.

In our hypothetical situations these networks (along with the original) could be stolen by a malevolent attacker, who would try to conceal his thievery by using the following three methods: adding additive noise proportionally to the network weights, continuing network training on arbitrary data and network knowledge distillation.  All reported datapoints are an average of 25 independent measurements.

\vspace{-2mm}
\subsubsection{Dependence on Additive Noise} 

\begin{figure}[!htp]
\centering
\subfigure{\includegraphics[width=3.6in,height=1.6in]{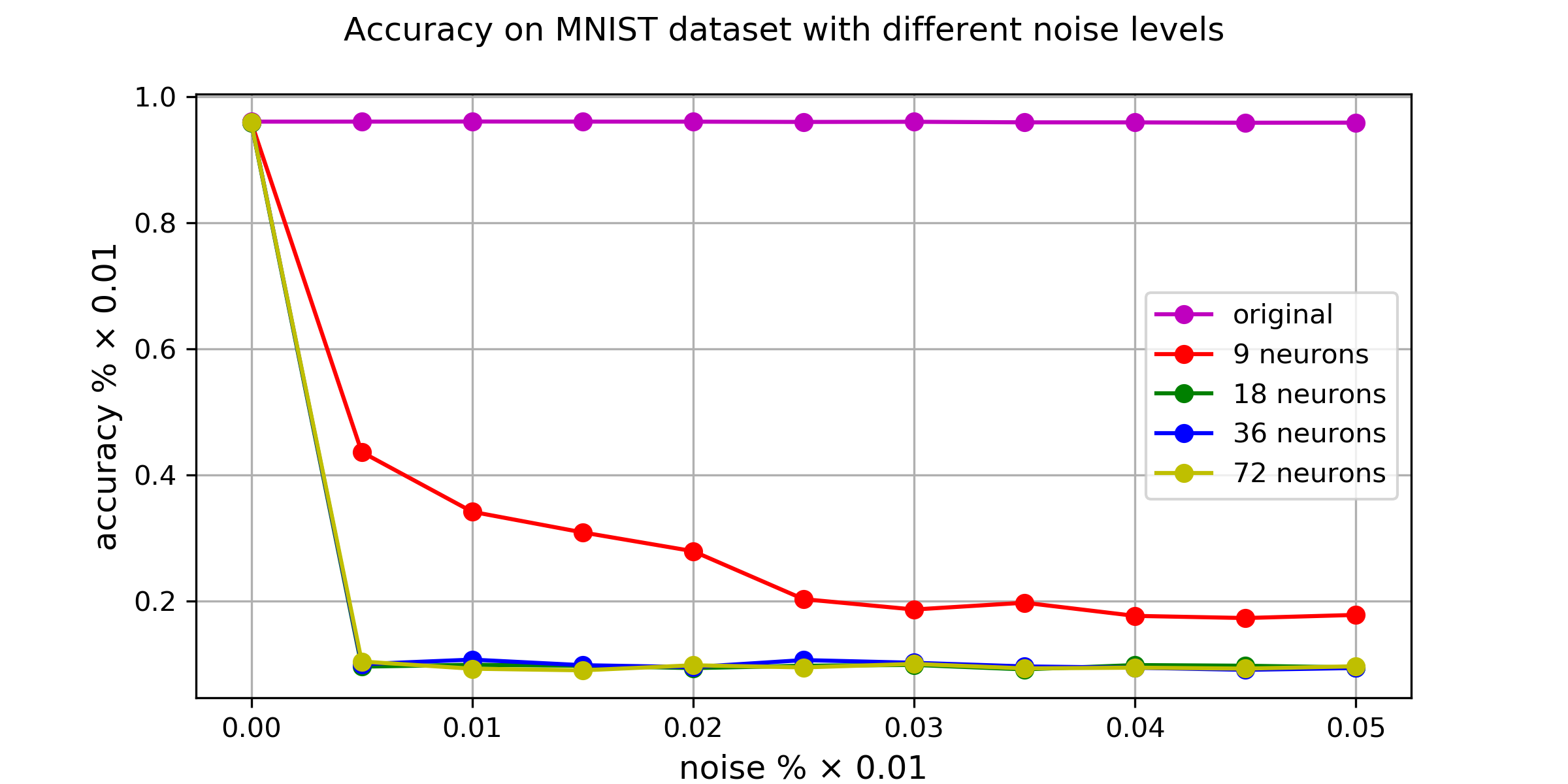}}
\caption{This figure depicts accuracy changes on the MNIST dataset under various level of additive noise applied on the weights. The original network (purple) is not dependent on these weight changes, meanwhile accuracies retrogress in the transformed networks, even with the lowest level of noise.}
\label{FigAddWeights}
\end{figure}

We have investigated network performance using additive noise to the network weights. The decrease of accuracy which depends on the ratio of the additive noise can be seen in Fig. \ref{FigAddWeights}.

At first we have tested a fully connected neural network trained on the MNIST dataset without making modifications to it. The decrease of accuracy was not more than 0.2\% even with a relatively high 5\% noise. This shows the robustness of a fully connected network.

\begin{figure}[!htp]
\centering
\subfigure{\includegraphics[width=3.6in,height=1.6in]{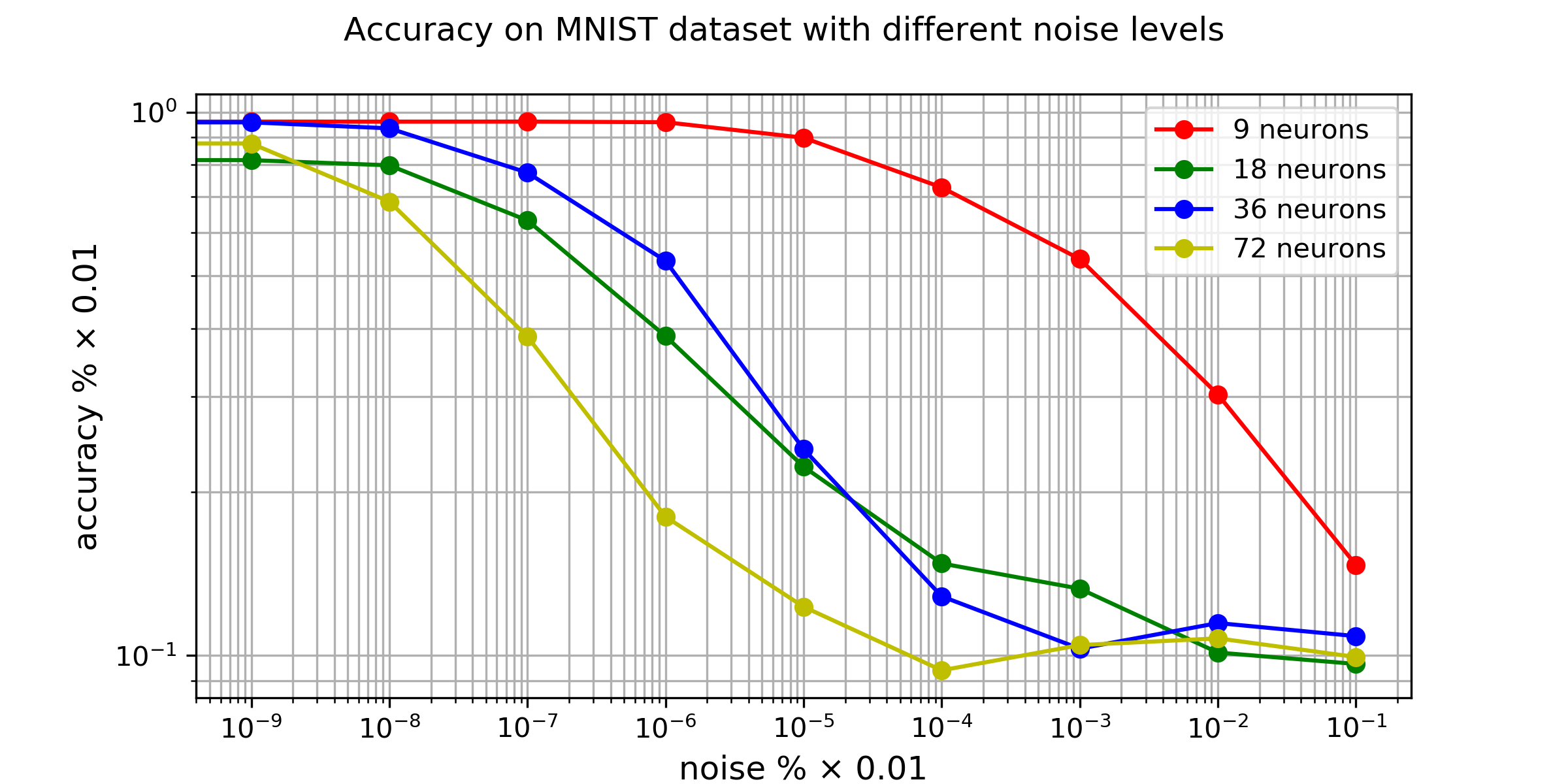}}
\caption{A logarithmic plot depicting the same accuracy dependence as on Fig. \ref{FigAddWeights}, focusing on low noise levels. As it can be seen from the plot, accuracy values do not change significantly under $10^{-7}$ percent of noise, which means the most important values of the weights would remain intact to proof connection between the original and modified networks. }
\label{FigAddWeightsLow}
\end{figure}

After applying the methods described in Sec. \ref{Transform} network accuracy retrogressed to $10\%$ even in case of noise which was less than $1\%$ of the network weights, as Fig. \ref{FigAddWeights} depicts.
This alone would reason the applicability of our method, but we have investigated low level noises further, which can be seen on Fig. \ref{FigAddWeightsLow}.
As it can be seen from the figure, accuracy starts to drop when the ratio of additive noise reaches the level of $10^{-7}$, which means the attacker can not significantly modify the weights. This effect could be increased by adding more and more neurons to the network.

\subsubsection{Dependence on Further Training Steps}
\begin{figure}[!htp]
\centering
\subfigure{\includegraphics[width=3.6in,height=1.6in]{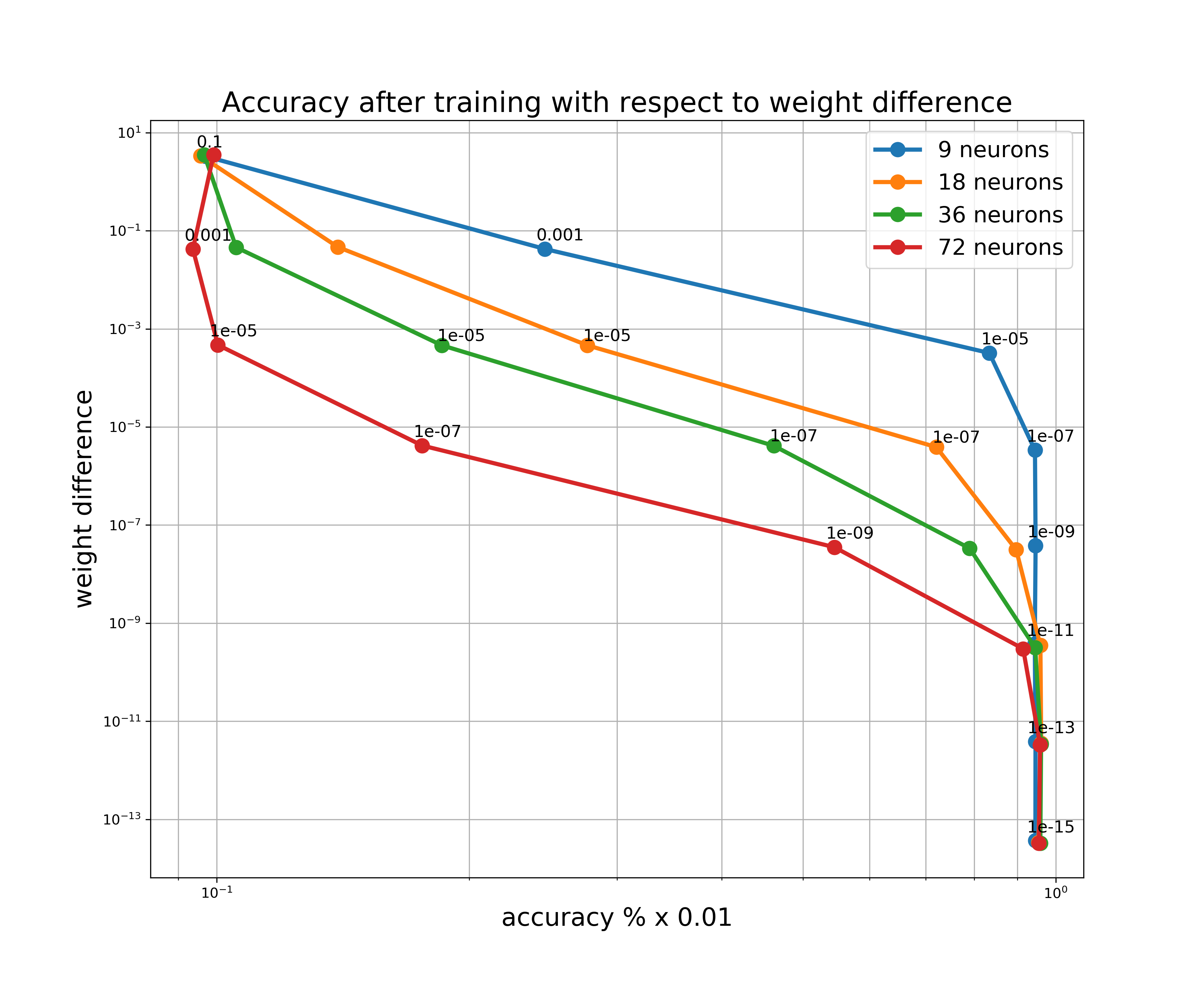}}
\caption{The figure plots accuracy dependence of the networks in case of further training (applying further optimization steps). As it can be seen from the plot weights had to be kept in $10^{-7}$ average distance to keep the same level of accuracy.} \label{trainingMNIST}
\end{figure}

Additive noise randomly modifies the weights, but it is important to examine how accuracy changes in case of structured changes exploiting the gradients of the network. Fig. \ref{trainingMNIST} depicts accuracy changes and average in weights distances by applying further training steps in the network.
Further training was examined using different step sizes and optimizers (SGD,AdaGrad and ADAM) training the network with original MNIST and randomly selected labels and the results were qualitatively the same in all cases. 

\subsubsection{Dependence on Network Distillation}

We have tried to distill the knowledge in the network and train a new neural network to approximate the functionality of previously selected layers, by applying the method described in \cite{hinton2015distilling}.

We have generated one million random input samples with their outputs for the modified networks and have used this dataset to approximate the functionality of the network.

We have created three-layered neural networks containing 32, 48, 64, 128 neurons in the hidden layer (The number of neurons in the first and last layer were determined by the original network) and tried to approximate the functionality of the hidden layers of the original structure. 
Since deceptive neurons have activations in the same order of magnitude as the original responses, these values disturb the manifold of the embedded representations learned by the network and it is more difficult to be approximated by a neural network. 
Table \ref{TableDistill} contains the maximum accuracies which could be reached with knowledge distillation, depending on the number of deceptive neurons and the neurons in the architecture used for distillation. 
This demonstrates, that our method is also resilient towards knowledge distillation.

\begin{table}[h!tb]
\begin{center}
\scalebox{1.0}{
\begin{tabular}{|| c || c| c| c| c||}
\hline
\#Dec. N. &  \#N. 32  & \#N. 48 & \#N. 64 & \#N. 128 \\
\hline\
9   & 0.64  & 0.65 & 0.69  & 0.71\\ 
\hline
18  & 0.12  & 0.14  & 0.15  & 0.17\\ 
\hline
36  & 0.10  & 0.11  & 0.10  & 0.13\\ 
\hline
72  &  0.11  & 0.09  & 0.10  & 0.10\\ 
\hline
\end{tabular}
}
\end{center}
\caption{The table displays the maximum accuracies reached with knowledge distillation, The different rows display the number of extra neurons which were added to the investigated layer, and the different columns show the number of neurons in the hidden layer of the fully connected architecture, which was used for distillation.}\label{TableDistill}
\end{table}


\vspace{-6.5mm}
\section{Conclusion}\label{Sec_Conclusion}
In this paper we have shown a transformation method which can significantly increase a network's dependence on its weights, keeping the original functionality intact.  We have also presented how deceptive neurons can be added to a network, without disturbing its original response. Using these transformations iteratively one can create and openly share a trained network, where it is computationally extensive to reverse engineer the original network architecture and embeddings in the hidden layers. The drawback of the method is the additional computational need for the extra neurons, but this is not significant, since computational increase is polynomial.

We have tested our method on simple toy problems and on the MNIST dataset using fully-connected neural networks and demonstrated that our approach results non-robust networks for the following perturbations: additive noise, application of further training steps and knowledge distillation.

\section*{Acknowledgements}

This research has been partially supported by the Hungarian Government by the following grant: 2018-1.2.1-NKP-00008: Exploring the Mathematical Foundations of Artificial Intelligence.


\end{document}